\documentclass[11pt, a4paper, copyright]{weco_paper}

\usepackage{wrapfig}
\usepackage{float}
\usepackage{algorithm}
\usepackage{algpseudocode}
\usepackage{cleveref}
\usepackage{comment}
\Crefname{algorithm}{Algorithm}{Algorithms}
\crefname{algorithm}{alg.}{algs.}

\newcommand{\aidek}[1]{\ensuremath{\mathrm{AIDE}_{#1}}}
\newcommand{\aidehuman}{\ensuremath{\mathrm{AIDE}_{\mathrm{human}}}}
\newcommand{\aidesq}{\ensuremath{\mathrm{AIDE}^{2}}}

\title{Recursive self-improvement of AI research agents}

\author[1]{Dhruv Srikanth}
\author[1]{Bingchen Zhao}
\author[1]{Dixing Xu}
\author[1]{Yuxiang Wu}
\author[1]{Zhengyao Jiang}

\affil[1]{Weco AI}

\correspondingauthor{dhruv@weco.ai, zhengyao@weco.ai}

\hypersetup{%
  pdftitle={Recursive self-improvement of AI research agents},%
  pdfauthor={Dhruv Srikanth, Bingchen Zhao, Dixing Xu, Yuxiang Wu, Zhengyao Jiang}%
}

\begin{abstract}
AI agents are beginning to automate research and development across the AI stack, from improving training efficiency to optimizing inference. A natural next step is to improve the research efficiency of the agents themselves.
When an AI research agent's own code is the object of optimization, each accepted rewrite becomes the agent that the next round edits. We refer to this loop as recursive self-improvement.
Its significance lies in a long-standing trend, in which increased cumulative spending on R\&D yields diminishing returns.
Sustained self-improvement offers a way to counter this trend.
We present \aidesq{}, a system that implements this loop for a frontier AI research agent.
It proposes changes to its own code, benchmarks modified versions of itself on a suite of AI R\&D tasks, and keeps the changes that perform best on hidden evaluations.
In an autonomous 8-day run, \aidesq{} discovered seven successive improvements, ranging from a new search policy to memory mechanisms that compress and manage the agent's growing context.
These gains generalize to four held-out benchmarks spanning machine learning engineering, heuristic algorithm engineering, and physics-based weather forecasting, the last of which is out of distribution from the selection tasks. On all four, the strongest discovered agent matches or exceeds a human-engineered production research agent that ranks among the strongest on FML-Bench.
On a separate held-out task family, the discovered agents also exhibit reduced reward hacking, a property the loop never explicitly optimized for: the rate falls from 55\% to 32\% during the run, 7 percentage points below the human-engineered agent.
Together, these results show that an AI research agent can improve its own research efficiency through recursive self-improvement, and that these gains transfer to tasks and domains the loop never encountered.
\end{abstract}

\begin{document}

\maketitle

\section{Introduction}
\label{sec:introduction}

AI agents are now used extensively to accelerate and automate parts of research and development (R\&D) across the AI stack, from machine learning engineering \citep{aide2025, airadojo, karpathy2026autoresearch} and GPU kernel engineering \citep{alphaevolve,kernelevolve2025} to algorithmic discovery \citep{liu2024llm4ad, shinkaevolve} and the design of agent pipelines and harnesses \citep{zhang2024aflow, agrawal2025gepa, adas, metaharness}.
Beyond individual components, agents now run complete research workflows, from generating research ideas \citep{si2024ideas, baek2025researchagent} to executing experiments and writing papers \citep{schmidgall-etal-2025-agent, aiscientist2026, jansen2025codescientist}.
Such systems improve the efficiency of the artifacts they produce, such as training and inference efficiency, yet the efficiency of the research process producing them remains fixed. In conventional R\&D, further progress requires increased human effort, making continued improvement increasingly costly as research becomes more difficult~\citep{bloom2020ideas}. Recursive self-improvement (RSI) points the research process at itself.
The prospect of an AI system that improves itself more efficiently than human researchers has been discussed since the earliest days of computing~\citep{good1965speculations} and remains an active subject of debate~\citep{yudkowsky2013iem, weston2025coimprovement}.
Recent self-improving systems show promise in coding and other domains~\citep{zelikman2023stop, dgm, hgm, hyperagents}, yet the effectiveness of their improvement loops at optimizing frontier AI research efficiency remains unclear.

\begin{figure}[t]
    \centering
    \includegraphics[width=\textwidth]{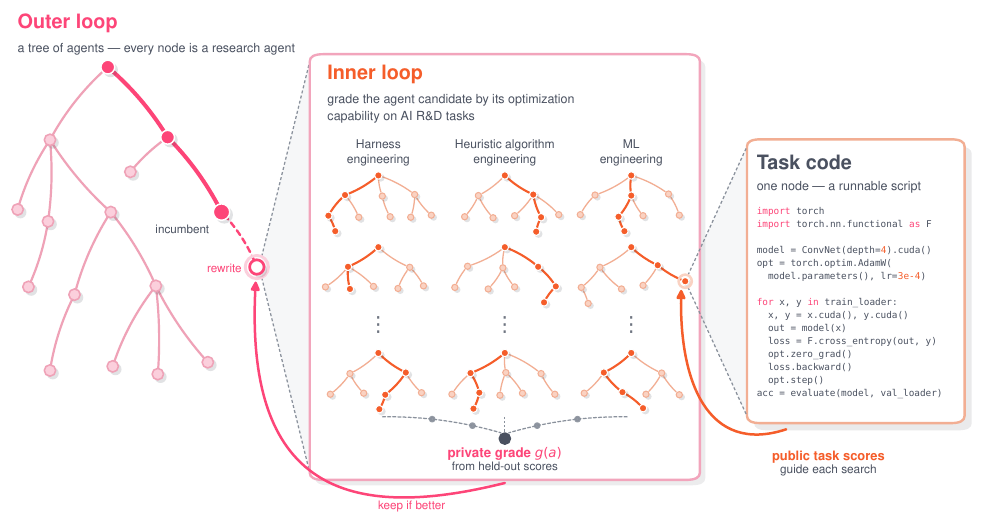}
    \caption{\textbf{One step of recursive self-improvement, shown at three levels of magnification.} Left: the outer loop searches a tree of agents, where every node is a complete research agent. The outer-loop agent proposes a rewrite of the current incumbent's code. Middle: the candidate is graded by running it as an optimizer on AI R\&D tasks. Held-out scores across the tasks aggregate into the private grade $g(a)$. The rewrite is kept only if $g(a)$ improves on the incumbent's grade. Right: each solution node is a runnable script for its task.}
    \label{fig:system}
\end{figure}

In this work, we study how recursive self-improvement can effectively advance frontier AI R\&D.
Our experiments target the harness layer, the code that surrounds a model and controls an agent's search, context, and verification.
As shown in \cref{app:harness-model}, a substantial share of an agent's realized capability is determined here.
We present \aidesq{}, a two-loop system that carries out recursive self-improvement at the harness layer.
In the inner loop, a research agent optimizes code against a measurable objective on problems drawn from a diverse set of AI R\&D tasks.
The outer loop performs a meta-level optimization over this inner-loop research process, rewriting the research agent with the objective of improving the research efficiency of the inner-loop agent.
\Cref{fig:system} illustrates this two-loop process structured as a nested tree search.

In one autonomous 8-day run, \aidesq{} discovered seven successive improvements, each accepted only after it improved results on held-out data that the agent being rewritten never observes. We compare the discovered agents against \aidehuman{}, a production research agent developed over two years of human-driven R\&D that ranks among the strongest on FML-Bench~\citep{zou2026fmlbench}. The strongest discovered agent matches or exceeds this baseline on four external benchmarks that never influenced the run. On a separate held-out task family, the reward hacking rate falls from 55\% to 32\%, below the 39\% of the human-engineered agent. When used as the outer-loop agent, a discovered agent continues to produce accepted improvements, though due to compounding noise across both loops and the prohibitive cost of running additional seeds, its performance in that role cannot be decisively distinguished from the strong baseline.

In summary, we present \aidesq{}, a recursive self-improvement system in which an AI research agent improves its own research efficiency.
Across the recursive self-improvement run, the loop repeatedly finds and accepts improvements under a fixed evaluation budget.
Those improvements generalize beyond the tasks used to select them, carry over to a domain the loop never encountered, and come with a reduction in reward hacking that was not part of the objective being optimized for.
Measured against a strong baseline developed over two years of human-driven R\&D, the discovered agents match or outperform it on the held-out benchmarks.

\FloatBarrier
\section{Method}
\label{sec:method}

We frame recursive self-improvement as a bi-level optimization problem consisting of two loops that iteratively optimize an agent against a measurable outcome.
The inner loop acts on specific tasks designed to capture an agent's ability to optimize code within a subset of AI R\&D domains. The outer loop operates on a meta-level, working to improve the inner-loop agent's optimization capability. Each inner-loop agent is evaluated at a fixed budget; therefore, improving the optimization capability directly improves the research efficiency of the inner-loop agent. Each accepted rewrite becomes the agent that the next step edits and evaluates. Both loops are driven by agents of the same tree-search design. \Cref{fig:system} illustrates this bi-level optimization process.

\subsection{Recursive self-improvement}
\label{sec:loop}

\noindent\textbf{The inner loop.} Given a codebase and a measurable metric, an inner-loop agent, denoted $a$, iteratively edits the code to improve performance under that metric. We write $x$ for a candidate solution and $r^{\mathrm{pub}}_t$ for the signal the agent observes on a task $t$. Let $x_0$ denote the task's existing codebase and $x_{<i}$ the edits produced so far. The agent repeatedly proposes candidates (\cref{eq:inner}) until a fixed dollar budget $b_t$ is spent, covering the agent's tokens and the cost of executing its solutions.
\begin{equation}
\label{eq:inner}
x_i = a\big(x_{<i},\; r^{\mathrm{pub}}_t(x_{<i})\big)
\end{equation}
The agent then returns one solution of its own choosing, which we denote $\hat{x}_t$. Which candidate to edit next, which to return, and how to structure the agent's memory are all part of the agent's code and are editable by the outer loop.

\noindent\textbf{Grading an agent.} When evaluated, an agent receives a fixed set of tasks spanning three families of AI R\&D work, which we refer to as the \textit{selection benchmark} (detailed in \cref{sec:system}). We construct the task set to be diverse in order to induce evolutionary pressure on the types of improvements that are made to the inner-loop agent, namely general mechanisms over task-specific tricks. Once the agent is run on each task, it returns a candidate solution to be scored on private held-out data using $r^{\mathrm{priv}}_t$. The resulting private scores are then aggregated:
\begin{equation}
\label{eq:grade}
g(a) = \frac{1}{T}\sum_{t=1}^{T} r^{\mathrm{priv}}_t\big(\hat{x}_t\big),
\end{equation}
where $\hat{x}_t$ is the solution returned by the agent $a$ on task $t$ and $T$ is the number of tasks in the selection benchmark. In practice, each task is run several times independently and the scores are averaged across runs.

\noindent\textbf{The outer loop.} An agent is itself written in code. Therefore, the same procedure can be applied one level up. At step $k$, the outer-loop agent $a^{\mathrm{out}}$ reads the previously proposed agents and their grades and proposes a rewrite. Under the substitution $(a,\, x,\, r^{\mathrm{pub}}_t) \mapsto (a^{\mathrm{out}},\, a,\, g)$, we can rewrite \cref{eq:inner} as:
\begin{align}
\label{eq:outer-propose}
a_k &= a^{\mathrm{out}}\big(a_{<k},\; g(a_{<k})\big)
\end{align}

In practice, the outer-loop agent edits the current incumbent, so each accepted rewrite becomes the codebase that is edited at the next step. At step $k$, the incumbent is the best agent graded so far. Candidate selection differs between the two levels: inner-loop selection is part of the agent's editable policy and is repeatedly rewritten during the run, whereas outer-loop selection is fixed by:
\begin{align}
\label{eq:outer-select}
a^*_k &= \underset{a \,\in\, a_{\le k}}{\arg\max}\; g(a)
\end{align}

Since the outer loop selects inner-loop agents according to $g$, two properties of $g$ shape the resulting selection pressure. First, separating the inner loop's optimization signal $r^{\mathrm{pub}}_t$ from the outer loop's selection signal $g$ (driven by $r^{\mathrm{priv}}_t$) prevents the inner-loop agent from directly optimizing the criterion used for outer-loop selection. Second, evaluating all agents under the same per-task budget $b_t$ constrains improvements to arise from a better algorithm rather than from spending additional compute. We describe the procedure for recursive self-improvement in \cref{algo:rsi}.

\begin{algorithm}[!htbp]
\caption{Recursive self-improvement}
\label{algo:rsi}
\begin{algorithmic}[1]
\Require initial agent $a_0$, outer-loop agent $a^{\mathrm{out}}$, tasks $\{t\}$ with budgets $\{b_t\}$, trajectory length $K$
\Function{Grade}{$a$}
  \ForAll{tasks $t = 1, \dots, T$} \Comment{run $a$ on each task}
    \State $x_0 \gets$ task $t$'s existing codebase; $i \gets 0$
    \While{total cost $< b_t$} \Comment{agent tokens + solution execution}
      \State $i \gets i+1$
      \State $x_i \gets a\big(x_{<i},\, r^{\mathrm{pub}}_t(x_{<i})\big)$ \Comment{propose the next solution}
    \EndWhile
    \State $\hat{x}_t \gets$ the solution $a$ selects from $x_{\le i}$
  \EndFor
  \State \Return $\frac{1}{T}\sum_{t=1}^{T} r^{\mathrm{priv}}_t\big(\hat{x}_t\big)$ \Comment{private held-out grading}
\EndFunction
\Statex
\State $g(a_0) \gets \Call{Grade}{a_0}$ \Comment{establish the baseline grade}
\For{$k = 1, \dots, K-1$}
  \State $a_k \gets a^{\mathrm{out}}\big(a_{<k},\, g(a_{<k})\big)$ \Comment{propose an agent rewrite}
  \State $g(a_k) \gets \Call{Grade}{a_k}$
\EndFor
\State \Return $a^*_{K-1} = \arg\max_{a \in a_{<K}} g(a)$
\end{algorithmic}
\end{algorithm}

\FloatBarrier
\subsection{\texorpdfstring{\boldmath\aidesq{}}{AIDE\textsuperscript{2}}}
\label{sec:system}

We instantiate \cref{algo:rsi} through our system \aidesq{}. The inner-loop agent $a_{0}$ starts from \aidek{0}, a pared-down refactor of AIDE \citep{aide2025}. AIDE was originally designed for ML engineering and performed well on MLE-Bench \citep{mlebench}. However, due to the diversity of tasks we run RSI on, we require a more general-purpose optimization agent. To this end, we remove the ML-specific machinery but maintain the same tree-search procedure. \aidek{0} grows a tree of candidate solutions rooted in an existing codebase, optimizing against a prespecified metric. The agent consists of several operators designed for different purposes: \textit{draft} to explore broad sets of new ideas, \textit{debug} to repair a solution with a bug or execution error, and \textit{improve} to refine a promising direction. Which solution is operated on is determined by the agent's search policy. In \aidek{0}, the selection is greedy: the highest-scoring solution becomes the parent for the next operation. \aidek{0} employs a reviewer agent that reads execution outputs from evaluating solutions and extracts a score and any relevant feedback.

The outer-loop agent $a^{\mathrm{out}}$ is driven by \aidehuman{}, an autonomous research agent used in production and developed by Weco's R\&D team. \aidehuman{} largely follows the same design choices and agent architecture as \aidek{0}. In addition to serving as the outer-loop agent, \aidehuman{} is used as a strong baseline developed through human-driven R\&D when testing generalization at various levels: on the selection benchmark (\cref{sec:run}), on held-out tasks in and out of distribution (\cref{sec:transfer}), and in the discovered agents' ability to act as a better self-improver (\cref{sec:ignition}). In \cref{app:fml}, we show that \aidehuman{} is competitive with existing code optimization agents, supporting its use as a strong baseline. During the recursive self-improvement run, we hold the model fixed within each loop. The outer-loop agent runs on \texttt{claude opus 4.7} \citep{opus-4-7}, while every inner-loop agent is evaluated with \texttt{gemini 3 flash} \citep{gemini-3-flash}. At the task-specific budgets $b_t$, \texttt{gemini 3 flash} matched or slightly exceeded the more expensive models tested on the selection tasks, so we use it for all inner-loop evaluations. Because evaluating each candidate dominates the cost of proposing it, we use the more capable model to drive the outer-loop agent. For the same reason, the inner-loop reviewer makes a single LLM call over each solution's execution output, while the outer-loop agent's reviewer explores the evaluation artifacts over several steps.

The selection benchmark consists of three task families. \emph{ML engineering} asks the agent to train a model against a target metric. \emph{Heuristic algorithm engineering} covers competitive-programming-style combinatorial problems where progress comes from iterating on algorithms and heuristics. \emph{Harness engineering} targets improvements to an agent's prompts, context management, and feedback loops---the code and algorithms that turn LLM API calls into working agentic systems. As shown in \cref{fig:system}, every inner-loop agent $a_k$ is run across the task families under a fixed budget and evaluated on a private held-out set. Its private scores are then aggregated to produce a grade $g(a_k)$, which \aidehuman{} uses as the selection signal to drive subsequent improvements of the inner-loop agent.

\FloatBarrier
\section{Experiments}
\label{sec:results}

\subsection{Experimental design}
\label{sec:design}

To better understand the effectiveness of recursive self-improvement on AI R\&D tasks, we use the following criteria and constraints when designing our experiments. The same run must produce several new incumbents throughout the optimization, as one favorable rewrite cannot show that the loop repeatedly finds useful improvements. In order to distinguish a sustained trend from a one-off gain, we test whether the recursive self-improvement run contains \textbf{repeated improvements}. We evaluate every agent produced during the recursive self-improvement run under a \textbf{fixed evaluation budget} so that a measured gain reflects a better algorithm rather than additional compute. At a fixed cost, a gain in optimization capability represents a gain in research efficiency. Improvements selected on a set of tasks must remain useful on tasks that were not involved in selection. We test for this \textbf{generalization} to rule out overfitting to the selection benchmark. Lastly, we evaluate the discovered agents against a \textbf{strong baseline}: \aidehuman{}, a production research agent developed through human-driven R\&D that performs competitively on FML-Bench \citep{zou2026fmlbench}, a benchmark of AI R\&D tasks (see \cref{app:fml}).

\subsection{A sustained trend of improvements}
\label{sec:run}

\begin{figure}[!htbp]
    \centering
    \includegraphics[width=\linewidth]{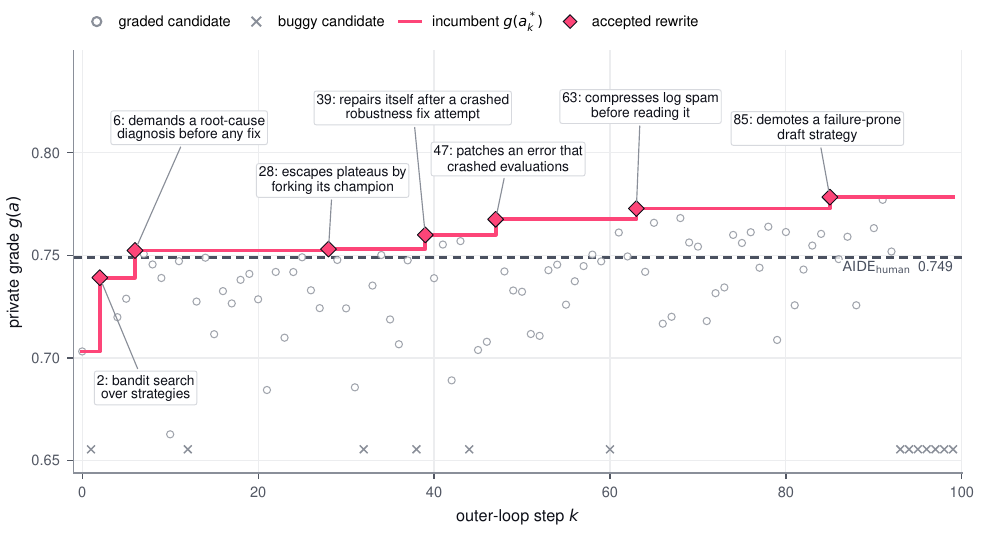}
    \caption{\textbf{A sustained trend of improvements.} Private grades $g(a_k)$ across the 100-node trajectory of an RSI run. Circles show graded candidates, crosses show buggy proposals, and diamonds mark the seven accepted rewrites that lift the incumbent grade $g(a^{*}_{k})$ from 0.703 to 0.778. The dashed line marks \aidehuman{} at 0.749 under the same grade. The vertical axis is truncated at 0.644. Buggy proposals receive no grade and are drawn in a row near the bottom edge. One early candidate scored 0.565 and falls outside the plotted range.}
    \label{fig:run-trace}
\end{figure}

We ran \aidesq{} over 8 days of wall-clock time, producing a 100-node trajectory, containing the initial agent and 99 rewrite proposals. As shown in \cref{fig:run-trace}, we observe seven accepted improvements at steps 2, 6, 28, 39, 47, 63, and 85, with the incumbent grade rising from 0.703 to 0.778, evidence of a sustained trend. Two further complete runs of the same protocol also produced sustained improvements, accepting two and four rewrites, respectively. Because candidates were selected on $g$, the recursive self-improvement trace is not meant to demonstrate generalization beyond the selection benchmark. However, given that the inner-loop agents use $r_t^{\mathrm{pub}}$ as their optimization signal and $g$ is a function of $r_t^{\mathrm{priv}}$, we use the grades to test each candidate agent's ability to produce solutions that generalize from the public signal to the private held-out data within each task. \Cref{fig:run-trace} shows that the improved agents eventually outperform \aidehuman{} on the selection benchmark; we refer to this as \textit{first-order generalization}. We test whether these gains transfer to held-out benchmarks in \cref{sec:transfer}.

\FloatBarrier
\subsection{Generalization to held-out benchmarks}
\label{sec:transfer}

The private grades used during recursive self-improvement cannot by themselves establish generalization beyond candidate selection. We instead test \textit{second-order generalization}: whether the improvements remain useful on four external benchmarks that never affected candidate selection.

\emph{ALE-Bench} \citep{imajuku2025alebench} evaluates agents on long-horizon combinatorial optimization from AtCoder programming contests. \emph{MLE-Bench} \citep{mlebench} evaluates agents on autonomous ML engineering across various Kaggle competitions. \emph{FML-Bench} \citep{zou2026fmlbench} evaluates agents on realistic research codebases, covering tasks from core ML problems including continual learning, causality, privacy, and robustness. These three benchmarks belong to task families represented during recursive self-improvement but have no overlap with the selection tasks, so we treat them as in-distribution at the task-family level. We use \emph{WeatherBench~2} \citep{weatherbench2} as the basis for a physics-based forecasting optimization task. Because neither weather forecasting nor physics-engine optimization is represented in the task set used for recursive self-improvement, we treat this as out-of-distribution. Additional details can be found in \cref{app:evaluation}.

We compare the agents \aidek{0}, \aidek{47}, \aidek{85}, and \aidehuman{} under a fixed set of per-benchmark constraints described in \cref{app:evaluation}. \aidek{0} is the baseline for measuring performance changes along the discovered lineage, while \aidehuman{} is the strong baseline developed through human-driven R\&D. \aidek{47} is the incumbent within the first 50 nodes, while \aidek{85} is the final incumbent from the recursive self-improvement trajectory described in \cref{sec:run}.

\begin{figure}[!htbp]
    \centering
    \includegraphics[width=\linewidth]{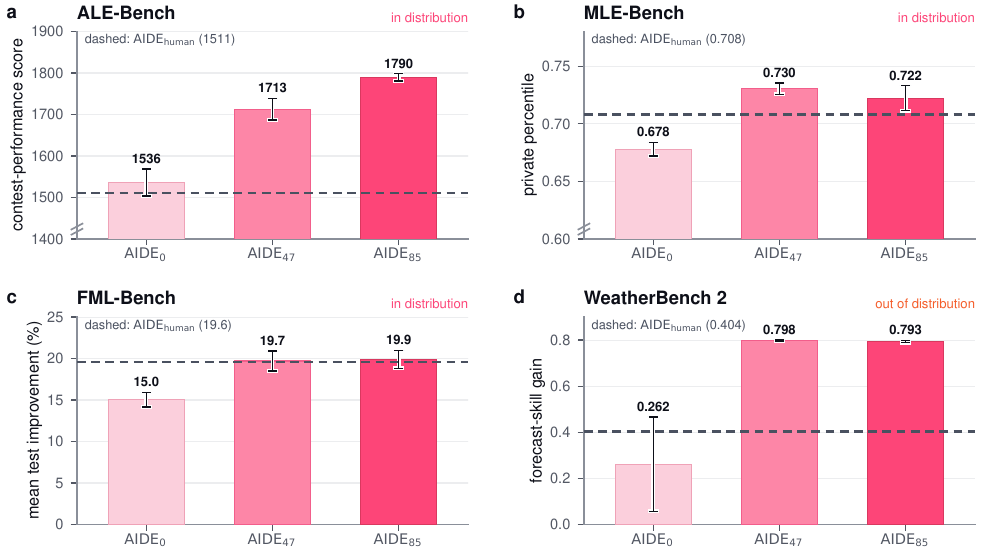}
    \caption{\textbf{Gains from recursive self-improvement transfer to held-out benchmarks.} \textbf{a}, Mean private contest performance on ALE-Bench (10 tasks, 10 seeds per task). \textbf{b}, Mean private percentile on MLE-Bench (22 tasks, 3 seeds per task). \textbf{c}, Mean normalized test improvement on FML-Bench (18 tasks, 3 seeds per task). \textbf{d}, Forecast-skill gain on WeatherBench~2 (one task, 3 seeds). Bars show benchmark means. Dashed lines mark \aidehuman{} under the same protocol. Error bars show $\pm$1 standard error of the benchmark mean. Higher is better. The axes in \textbf{a} and \textbf{b} are truncated, as indicated by the break marks. Further details are in \cref{app:evaluation}.}
    \label{fig:transfer}
\end{figure}

As shown in \cref{fig:transfer}, both evolved checkpoints improve on \aidek{0}, and \aidek{85} matches or exceeds \aidehuman{} on all four external benchmarks. The gains are positive throughout but not monotone across checkpoints: \aidek{85} performs best on ALE-Bench and FML-Bench, whereas \aidek{47} performs best on MLE-Bench and WeatherBench~2. Because candidate selection during the recursive self-improvement run aggregates performance over a heterogeneous selection benchmark, some non-monotonicity among strong checkpoints is expected. Nevertheless, the gains transfer from the recursive self-improvement run to benchmarks that are external to the run.

Interestingly, some of the largest performance gains appear on the out-of-distribution benchmark, WeatherBench~2, where the agents optimize the core of a physics-based weather-forecasting model, a domain and type of problem absent from the selection tasks. On every seed, both evolved checkpoints independently converged on the same family of changes to the forecasting model's numerics, reaching nearly identical gains with almost no variation across runs. \aidek{0} and \aidehuman{} reach a comparable solution on at most one seed and vary widely across the rest. This suggests that the accepted rewrites improve optimization behavior that is general enough to extend to an unfamiliar scientific-computing domain.

In \cref{sec:run,sec:transfer}, we demonstrate improvements concerning the agents' core capability of optimizing code against a specified metric. \Cref{sec:hacking} examines how recursive self-improvement affects the discovered agents' behavior on objectives the agents were \emph{not explicitly optimizing for}.

\FloatBarrier
\subsection{An emergent behavior: reduced reward hacking}
\label{sec:hacking}

\begin{wrapfigure}{r}{0.50\textwidth}
    \vspace{-1.0\baselineskip}
    \centering
    \includegraphics[width=\linewidth]{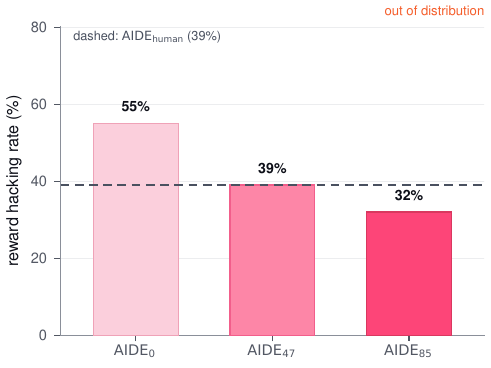}
    \caption{\textbf{Reward hacking declines along the discovered lineage} (lower is better). Bars show the percentage of 38 held-out (kernel, training-context) pairs on which each discovered agent reward hacks. The dashed line marks \aidehuman{} at 39\% under the same protocol. \Cref{app:evaluation} describes how the pairs are scored.}
    \label{fig:hacking-rate}
    \vspace{-0.8\baselineskip}
\end{wrapfigure}

As \cref{algo:rsi} selects agents by a numeric grade, the discovered agents could plausibly favor mechanisms that inflate their scores without corresponding downstream gains. As discussed in \cref{sec:loop}, we design \cref{algo:rsi} to limit this risk by decoupling the inner- and outer-loop optimization signals. In addition, the generalization results in \cref{sec:transfer} show no indication that the improvements overfit or exploit the selection grade. To examine this further, we measure the discovered agents' reward hacking rate on kernel engineering tasks, a task family not in the selection benchmark.

Reward hacking occurs when optimizing an imperfect proxy improves measured performance without a corresponding improvement in the intended objective \citep{krakovna2020specification,skalse2022defining}. Recent coding-agent studies measure this divergence by comparing agent-visible feedback with held-out outcomes, and find that the gap widens in longer-horizon and more complex tasks \citep{zhao2026specbench}. We follow the proxy-to-downstream design applied to GPU kernel engineering in \citet{zhao2026specbench} and evaluate agents on a subset of KernelBench \citep{ouyang2025kernelbench}, a benchmark in which agents replace PyTorch reference modules with custom GPU kernels, scored on numerical correctness against the reference and on wall-clock speedup over it. Each agent optimizes kernels for an isolated speedup measurement---the proxy it observes---and we then insert its kernels into GPT-2 \citep{radford2019language}, ViT \citep{dosovitskiy2021image}, and CNN \citep{lecun1998gradient} training loops to measure how much of the proxy gain survives. As shown in \cref{fig:hacking-rate}, the measured reward hacking rates decline along the discovered lineage: 55\% for \aidek{0}, 39\% for \aidek{47}, and 32\% for \aidek{85}, compared to 39\% for \aidehuman{}. These reward hacking rates establish a held-out behavioral change, though they do not identify which rewrites produced it. As recursive self-improvement progresses, cumulative harness rewrites selected on a grade shift a behavior that the grade never measured, specifically the agents' tendency to reward hack.

\FloatBarrier
\subsection{\texorpdfstring{\boldmath The self-improved agent: \aidek{85}}{The self-improved agent: AIDE85}}

\label{sec:discoveries}

In this section, we examine \aidek{85}, the agent discovered through recursive self-improvement. Two properties of \aidek{0} are relevant to the changes \aidek{85} introduces. First, \aidek{0} selects greedily, building on the highest-scoring candidate, with no explicit control over the trade-off between exploring new approaches and refining the current best. Second, it performs minimal history compaction---each drafting and improvement prompt receives the full concatenated history of prior candidates and their execution output, so prompts grow as the run proceeds. \aidek{85} redesigns both components and adds robustness mechanisms that may account for the reduced reward hacking observed in \cref{sec:hacking}.

\noindent \textbf{Search policy.} \aidek{85} employs bandit selection over drafting strategies with periodic forking.
Where \aidek{0} greedily improves the highest-scoring candidate with little control over when to explore, \aidek{85} runs a bandit policy \citep{auer2002bandit} over drafting \emph{strategies}.
Drafts are generated under one of five fixed strategies (\textit{conservative}, \textit{aggressive\_rewrite}, \textit{ensemble}, \textit{tuned\_specialist}, \textit{robust\_simple}), and each new node inherits the label from its parent.
At each step a strategy arm is chosen by UCB1, with 30\% of steps instead sampling an arm by a softmax over the arms' best scores. The highest-scoring node carrying the chosen arm's label is then expanded with a new node.
Selecting over strategy arms rather than individual nodes gives the agent a lever on the diversity of approach, not just on which local optimum to polish.

\begin{figure}[!htbp]
    \centering
    \includegraphics[width=\linewidth]{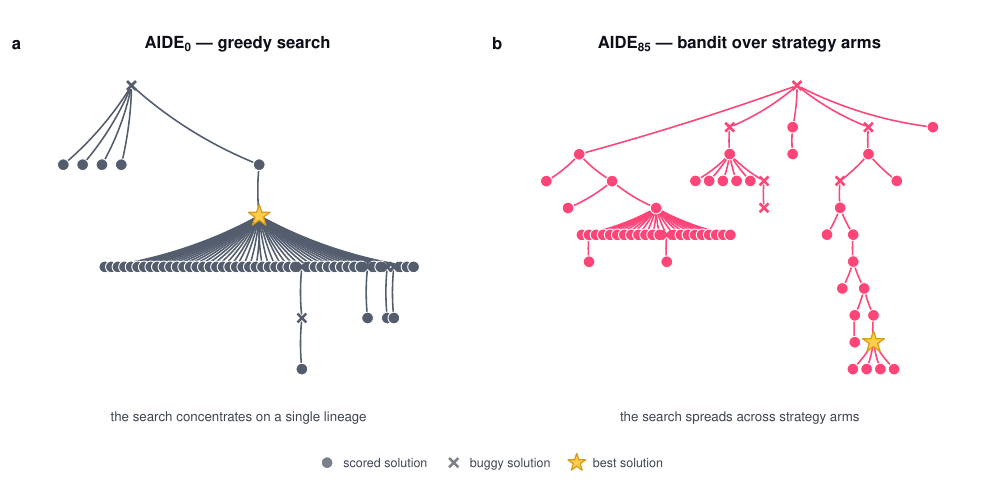}
    \caption{\textbf{Two search policies allocate the same budget differently.} Both agents run the same held-out ALE-Bench task, with the first 60 nodes shown. \textbf{a}, \aidek{0} improves the highest-scoring candidate, so nearly every solution descends from one draft. \textbf{b}, \aidek{85} keeps several arms alive, balancing exploration and refinement.}
    \label{fig:search-policy}
\end{figure}

\aidek{85} is also designed to handle stagnation of performance in the optimization process. Every five search steps, \aidek{85} forks the global best node, improving it under a different strategy arm.
\aidesq{} discovered that when the leading strategy plateaus, further refinement within that lineage yields diminishing returns, but restarting from a fresh draft would discard the strong solution already found. Forking avoids both failure modes by continuing to improve the current best node under a new strategy arm, and the five-step bound keeps this from interrupting otherwise productive refinement.
\Cref{fig:search-policy} shows how the two policies spend the same budget on a held-out task.

\begin{figure}[!htbp]
    \centering
    \includegraphics[width=\linewidth]{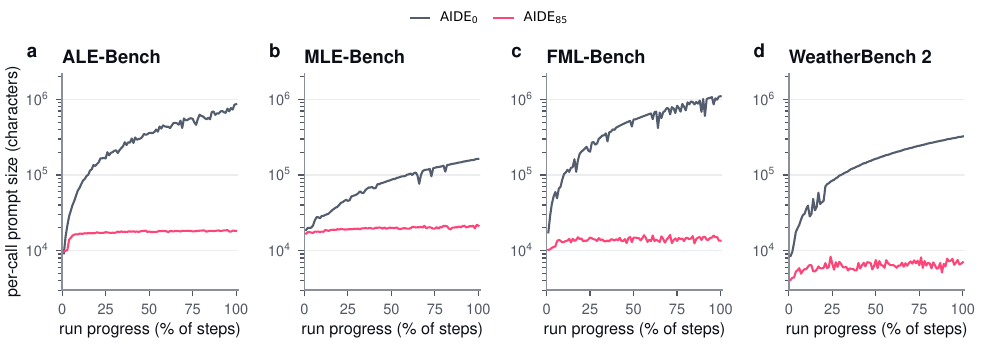}
    \caption{\textbf{The discovered agent holds per-call prompts to a bounded size.} Per LLM call prompt size in characters (log scale) across held-out runs on ALE-Bench, MLE-Bench, FML-Bench, and WeatherBench~2. Lines show the median over task means. \aidek{0}'s prompts grow with run history, while \aidek{85}'s stay roughly constant. A broader comparison between \aidek{0}, \aidek{47}, \aidek{85}, and \aidehuman{} can be found in \cref{fig:token-efficiency}.}
    \label{fig:token-efficiency-main}
\end{figure}

\noindent \textbf{Context management.} \aidek{85} uses bounded and role-specific prompts with a bug-rate-gated failure memory to compress and manage the agent's growing context.
In \aidek{0}, every drafting and improvement prompt contains the full history, so the prompt size grows with the search.
\aidek{85} instead has the \textit{draft} and \textit{improve} operators read a compact summary of the root and recent candidates rather than the full history.
\aidek{85} also uses a form of \emph{failure memory} in the context. When a run's candidates show a bug rate of at least 15\%, the agent injects up to three recurring error signatures (the final error lines of recent buggy candidates, deduplicated) into \textit{draft} and \textit{improve} prompts.
This 15\% threshold is a design choice from \aidesq{}. The mechanism stays dormant where bugs are rare and only activates when evaluation errors are frequent enough.
These mechanisms keep per-step prompt size roughly constant while \aidek{0}'s prompts grow with history.
The median task-level reduction in per-LLM-call prompt size compounds over a run, reaching 7$\times$ on MLE-Bench, over 40$\times$ on WeatherBench~2, and about 50$\times$ on ALE-Bench and FML-Bench (\cref{fig:token-efficiency-main}; see \cref{app:compression} for additional details).
Under fixed per-run constraints, shorter prompts let the agent make more effective use of each step and buy more search steps in cost-bound runs.

\noindent \textbf{Robustness.} \aidek{85} adds three robustness mechanisms.
The first is a pair of prompt-level safeguards: a fixed instruction in every code-generation prompt reminding the model that solutions are scored on a private split it cannot see and that it should prefer robust, generalizable approaches, and a guard that re-prompts when generated code is nearly empty (under 40 characters, e.g.\ a stub or placeholder).
The second is a selection rule meant to avoid picking a lucky one-off high score, penalizing each candidate by its distance from the median of the top candidates. Because the penalty preserves the ordering of the candidates, replaying the rule over the held-out runs shows that it never changed which candidate the agent selected. The most substantive change fixes a flaw in the evaluation itself rather than in the agent.
One task's held-out scoring script crashed on all of its test cases whenever any single test case failed, and \aidek{85} adds a small patch, which the agent described as ``\textit{a narrow, low-risk intervention targeting a verified failure mode \ldots{} without altering search dynamics}'', that stops a single failed test case from taking down the whole evaluation. Interestingly, rather than exploiting the broken evaluation, \aidesq{} repaired it.

\FloatBarrier
\subsection{Ignition test}
\label{sec:ignition}

For the gains in recursive self-improvement to turn diminishing returns into accelerating ones, we hypothesize that the discovered agents must be better at driving recursive self-improvement than the agent that discovered them. We call this comparison the \textit{ignition test}. A self-improved agent should be promoted to the outer loop only if it drives the loop more effectively than the agent that produced it. We run the ignition test using the incumbent agent after 50 outer-loop steps from the run described in \cref{sec:run}. The test comprises two independent arms of recursive self-improvement, each with three seeds run for 50 steps. Both arms start from the same inner-loop agent, \aidek{47}, and differ only in the agent used in the outer loop. The treatment arm uses \aidek{47}, the agent being tested, while the reference arm uses \aidehuman{} in the outer loop.

\begin{figure}[!htbp]
    \centering
    \includegraphics[width=0.84\linewidth]{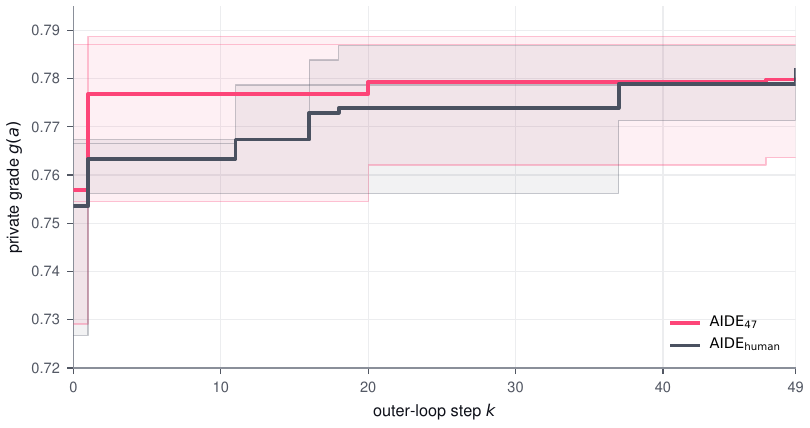}
    \caption{\textbf{The ignition test.} Both arms complete three independent 50-step runs, starting from the same inner-loop agent, \aidek{47}. Lines trace the incumbent private grade $g(a^{*}_{k})$. Line color identifies the outer-loop agent driving recursive self-improvement: pink for \aidek{47}, gray for \aidehuman{}. Thin lines show individual runs, shaded bands span the min--max range within each arm, and bold lines show each arm's average, ending at 0.780 and 0.782 for \aidek{47} and \aidehuman{} respectively.}
    \label{fig:ignition}
\end{figure}

Averaged across seeds, the two arms reach similar mean endpoints in the outer loop, with the reference arm finishing slightly higher, as shown in \cref{fig:ignition}. The treatment mean reaches its final score region after roughly 20 steps, compared with roughly 40 steps for the reference. However, with only three seeds per outer-loop agent, we find these results to be inconclusive; they do not establish that \aidek{47} is more sample-efficient as a self-improver than \aidehuman{}. Given the variance in final performance across seeds, we also do not claim that either agent is better than the other at driving recursive self-improvement. What the treatment arm does show is that there is no obvious degradation in \aidek{47}'s ability to drive the recursive self-improvement run compared to \aidehuman{}. Stronger conclusions are prohibitively costly as they would require additional outer-loop seeds and the evaluation of each seed's final agent on benchmarks external to the selection benchmark, following the protocol used in \cref{sec:transfer}.

\FloatBarrier
\section{Related work}
\label{sec:related}

\noindent\textbf{LLM-driven search.} As LLM capabilities improve, a growing body of work has focused on LLM-driven optimization loops that improve a candidate solution against a measurable objective \citep{lehman2022elm, liu2024llm4ad, airadojo, kernelevolve2025, cemri2026adaevolve, aira2}.
These loops differ mainly in how they structure the search. The simplest approach repeatedly edits a single artifact, keeping each edit that improves a validation metric \citep{karpathy2026autoresearch}.
AIDE \citep{aide2025} structures this as a tree search over candidate scripts, where each node represents a candidate solution and each edge a mutation between solutions. Building on ideas from genetic programming and quality-diversity search \citep{koza1994genetic, mouret2015mapelites}, several approaches widen this search structure to entire populations of programs \citep{romeroparedes2024funsearch,alphaevolve,openevolve,shinkaevolve,ray2026adaptevolve}.
These systems have yielded impressive improvements from discovering preference-optimization losses \citep{discopop} to generating an archive of diverse adversarial prompts against a fixed target \citep{rainbowteaming}.
Such open-ended generation of artifacts has been argued to be a prerequisite for reaching superhuman capability \citep{hughes2024openendedness}.
These systems optimize an artifact that aims to solve a given problem. The procedure that drives the optimization remains largely hand-designed through human-driven R\&D.

\noindent\textbf{Automated research.} A natural application of such search is the research process itself.
The AI Scientist \citep{aiscientist2026,aiscientistv2} automates ideation, experimentation, writing, and review in a single pipeline.
Related systems automate individual stages in the research pipeline, from idea generation to experiment execution \citep{baek2025researchagent, schmidgall-etal-2025-agent, jansen2025codescientist, researchpreferencemodels2026, murphy2026mda}.
Empirical accounts of such pipelines catalog recurring failure modes, from implementation drift to premature declarations of success \citep{trehan2026scientists}.
In the natural sciences, such systems have produced hypotheses and experimental protocols later validated in the laboratory \citep{gottweis2025coscientist, ghareeb2026multiagent, schmidgall2026coscientist}.
LLM-driven systems have also entered the domain of LLM development, including training data generation \citep{kulikov2026autodata}, pre-training data selection \citep{meng2026autodata}, environment engineering for continual learning \citep{liu2026spade}, and improving the pretraining procedure itself \citep{tan2026pretraining}.
Studies of these automated research pipelines find that LLM-generated research ideas can be judged more novel than those of expert humans \citep{si2024ideas}, though the advantage narrows once ideas are executed rather than only reviewed \citep{si2025executiongap}.
This gap motivates grounding automated research in execution outcomes \citep{si2026executiongrounded}.
A parallel effort measures how much of the research workflow agents can carry out, from replicating published papers to research engineering compared to human experts \citep{huang2023mlagentbench, wijk2024rebench, nathani2025mlgym, starace2025paperbench, zhao2025speedrun, airsbench2026, falck2026replica}.
Recent work also stresses guarding such benchmarks against exploitable shortcuts \citep{robustkbench2025} and separating the tasks used to develop a method from those used to evaluate it \citep{discogen2026}.

\noindent\textbf{Automated harness search.}
Some works focus on applying the same automated loop at the harness layer, studying whether this can yield better prompts, parameters, or workflows for entire agent pipelines \citep{zhou2023ape, pryzant2023protegi, yang2023opro, guo2023evoprompt, khattab2024dspy, wang2024promptagent, cheng2024trace, zhang2024aflow, agrawal2026optimizeanything}.
GEPA \citep{agrawal2025gepa} demonstrates that prompt evolution can outperform reinforcement learning.
TextGrad \citep{yuksekgonul2024textgrad} draws an analogy to gradient descent and treats text components as parameters, updating them by backpropagating natural-language feedback in place of gradients.
Some works extend the optimization target to the entire harness design, so the loop directly yields an agent ready for downstream tasks \citep{adas, metaharness}.
Harness designs differ along several axes, such as whether the agent accumulates reflections or skills as it acts \citep{shinn2023reflexion, wang2023voyager} or jointly adapts model weights with prompts or broader harness code \citep{tiwari2026fastslow, kim2026whale}.
In these works, the harness being optimized is separate from the procedure that optimizes it.

\noindent\textbf{Self-referential improvement.}
Self-referential improvement refers to when the optimization machinery itself becomes an object of optimization \citep{schmidhuber1993selfreferential, yang2026selfimprovementsurvey, gao2026selfevolvingsurvey, lin2026agenticevolution}.
The G\"odel machine \citep{godelmachine} permits any self-rewrite that provably yields higher expected utility than not making it.
Because such proofs are generally impractical, empirical successors substitute evaluation for proof. Some optimizers evolve their own mutation prompts or search strategies \citep{fernando2024promptbreeder, liu2026evox} while others let a single agent rewrite its own code, from an improver scaffold \citep{zelikman2023stop} to a full agent codebase \citep{yin2025godelagent, robeyns2025sica}, keeping the changes that improve task performance.
The Darwin G\"odel Machine \citep{dgm} keeps an archive of such self-rewriting coding agents. Subsequent work select ancestors by the aggregate success of their descendants \citep{hgm} or let rewrites reach the routine that proposes them \citep{hyperagents}.
Evaluation signals can be noisy or misleading, so related work strengthens the evidence required to accept a modification by gating each one behind a statistical measure \citep{statisticalgodelmachine2025} or co-evolving the evaluators with the agents they score \citep{rqgm2026redqueen}.
These works typically use a proxy metric to steer self-modification. Agent performance on software engineering, for example, can serve as a proxy for the ability to write code that improves the agent's own architecture \citep{dgm}.
\aidesq{} instead targets the agent's own capability on AI R\&D tasks.

\noindent\textbf{Meta-learning and learned optimizers.} Meta-learning seeks to improve a learning procedure using experience across tasks. Early work represented learning algorithms with recurrent neural networks \citep{hochreiter2001learning}, an approach developed further in neural optimizers that learn parameter updates \citep{andrychowicz2016learning} and black-box search strategies \citep{chen2017learningwithout}. Related approaches learn model initializations \citep{finn2017maml} or training hyperparameters \citep{maclaurin2015reversible} from performance after adaptation. These settings can be formulated as bi-level optimization \citep{franceschi2018bilevel}, with GIMLI providing a general framework for differentiable inner-loop meta-learning and \texttt{higher} supporting its implementation \citep{grefenstette2019generalized}. In reinforcement learning, meta-learned update rules have also yielded algorithms that transfer to environments outside those used for discovery \citep{oh2020discovering, oh2025discorl}.

Alongside these neural approaches, AutoML-Zero \citep{real2020automlzero} and the work introducing Lion \citep{chen2023symbolic} search directly over programs that implement learning algorithms and parameter-update rules. \aidesq{} applies this broader idea of optimizing an optimization procedure to AI research agents, searching over harness code and evaluating candidates by their downstream research performance under fixed budgets.

\FloatBarrier
\section{Discussion}
\label{sec:discussion}

\aidesq{} demonstrates that recursive self-improvement at the harness layer can produce transferable gains in an AI research agent's research efficiency. During the recursive self-improvement run, the loop accepted seven rewrites, each under a fixed evaluation budget (\cref{sec:run}). Under this fixed evaluation budget, gains in optimization capability on AI R\&D tasks translate to gains in research efficiency. The accepted rewrites concentrate on problems that practitioners face when building effective agentic systems: recovering from search plateaus, managing context under fixed budgets, and guarding against untrustworthy wins (\cref{sec:discoveries}). On four held-out benchmarks spanning in- and out-of-distribution tasks, \aidek{85} equals or surpasses \aidehuman{} (\cref{sec:transfer}), a strong baseline developed through human-driven R\&D (\cref{app:fml}). On a separate held-out task family, the reward hacking rate fell from 55\% to 32\%, a property the loop never explicitly optimized for (\cref{sec:hacking}). Such improvements have traditionally drawn on human-driven R\&D through engineering time, expertise, and domain understanding. However, a loop that proposes and validates potential improvements autonomously makes the harness layer amenable to search and learning, shifting part of the bottleneck from expert engineering effort toward compute.

Noise compounds across both loops of the bi-level optimization process. The trajectory of the inner-loop search varies across seeds, and the evaluation of any given solution from the trajectory can itself be noisy. The private grade $g(a)$ that determines the acceptance of a candidate rewrite reflects both sources. If the noise is high enough, a falsely accepted rewrite becomes the new incumbent (\cref{eq:outer-select}), so a single noisy comparison can derail the outer loop's subsequent search. This same noise limits the conclusiveness of the ignition test (\cref{sec:ignition}). The results for \aidek{47} show a possible gain in sample efficiency as the outer-loop agent without obvious degradation in performance after 50 steps relative to the \aidehuman{} reference arm. However, a definitive comparison would be prohibitively costly, requiring additional seeds of recursive self-improvement and a full held-out evaluation of each seed's final agent. This makes cost and access to compute limiting factors when evaluating such systems.

Despite the gains across various benchmarks, the discovered agents present practical challenges. They remain complex and difficult to interpret. It is unclear which components drive performance and which, if any, are unused artifacts of earlier recursive self-improvement steps. This complexity can increase deployment friction due to the requirements of production systems such as maintaining compatibility with existing product features and adhering to the constraints of the deployment infrastructure.

\section*{Acknowledgments}
We thank Jean Kaddour, Minqi Jiang, Morgan McGuire, George Zhang, Ross Taylor, Ofir Press and Christian Schwarz for their helpful feedback.
We also thank Qiran Zou for sharing the FML-Bench results used in our comparison against other code optimization agents.

{\small
\bibliography{paper}
}

\clearpage
\appendix

{\centering \LARGE \textbf{Appendix}}

\section{Held-out benchmark details}
\label{app:evaluation}

In \cref{sec:transfer}, we evaluate the four agents \aidek{0}, \aidek{47}, \aidek{85}, and \aidehuman{} across various benchmarks external to the selection benchmark. The held-out benchmark scores from \cref{fig:transfer} are shown in \cref{tab:transfer-details}.
Within each comparison, all agents run under the same fixed protocol, the details of which are shown in \cref{tab:transfer-protocol}. We use a per-run cost cap on ALE-Bench, MLE-Bench, and WeatherBench~2, and a per-run step cap on FML-Bench and KernelBench (used to measure reward hacking in \cref{sec:hacking}) as the primary constraints. Additional constraints listed in \cref{tab:transfer-protocol} are safeguards that keep a run from continuing unbounded and are selected such that in most cases, they are not the binding constraint. A run that reaches a non-primary constraint, for example, the 12 hour time limit for ALE-Bench which uses cost as the primary constraint, is scored on the best candidate available at that point. When evaluating \aidek{0}, 5 FML-Bench runs and 48 ALE-Bench larger budget runs from \cref{app:harness-model} terminated early due to errors related to context-window limits of the underlying LLM. These runs were also scored on the best candidate available at termination.
ALE-Bench solutions run on 128-vCPU x86\_64 machines with 512 GB of RAM. MLE-Bench solutions are executed on CPU-only x86\_64 containers on Modal, from 4 vCPU with 16 GB of memory to 24 vCPU with 192 GB depending on the size of the dataset. WeatherBench~2, FML-Bench, and KernelBench use NVIDIA A100 GPUs to evaluate solutions.

\begin{table}[H]
    \centering
    \footnotesize
    \caption{\textbf{Held-out benchmark results.} We measure ALE-Bench using mean private contest performance, MLE-Bench using mean private percentile, WeatherBench~2 using forecast-skill gain, and FML-Bench using mean normalized test improvement, and report one standard error of the benchmark mean. The best score in each column is in bold.}
    \label{tab:transfer-details}
    \begin{tabular}{lcccc}
        \toprule
        \textbf{Agent} & \textbf{ALE-Bench} & \textbf{MLE-Bench} & \textbf{WeatherBench~2} & \textbf{FML-Bench (\%)} \\
        \midrule
        \aidek{0} & $1536 \pm 33$ & $0.678 \pm 0.006$ & $0.262 \pm 0.205$ & $15.0 \pm 0.9$ \\
        \aidek{47} & $1713 \pm 26$ & $\mathbf{0.730 \pm 0.005}$ & $\mathbf{0.798 \pm 0.003}$ & $19.7 \pm 1.2$ \\
        \aidek{85} & $\mathbf{1790 \pm 9}$ & $0.722 \pm 0.011$ & $0.793 \pm 0.005$ & $\mathbf{19.9 \pm 1.1}$ \\
        \aidehuman{} & $1511 \pm 35$ & $0.708 \pm 0.007$ & $0.404 \pm 0.193$ & $19.6 \pm 1.0$ \\
        \bottomrule
    \end{tabular}
\end{table}

\begin{table}[H]
    \centering
    \footnotesize
    \caption{\textbf{Evaluation protocols per task within each benchmark.} }
    \label{tab:transfer-protocol}
    \begin{tabular}{lcccccl}
        \toprule
        \textbf{Benchmark} & \textbf{Tasks} & \textbf{Seeds} & \multicolumn{3}{c}{\textbf{Constraint}} & \textbf{Model} \\
        \cmidrule(lr){4-6}
         & & & \textbf{Cost (\$)} & \textbf{Steps} & \textbf{Time (h)} & \\
        \midrule
        ALE-Bench (lite) & 10 & 10 & 5 & 200 & 12 & \texttt{gemini 3 flash} \citep{gemini-3-flash} \\
        MLE-Bench (lite) & 22 & 3 & 5 & 500 & 24 & \texttt{gemini 3 flash} \citep{gemini-3-flash} \\
        WeatherBench~2 & 1 & 3 & 15 & 200 & -- & \texttt{gemini 3.1 pro} \citep{gemini-3-1-pro} \\
        FML-Bench & 18 & 3 & -- & 100 & -- & \texttt{gpt-5.4} \citep{gpt-5-4} \\
        KernelBench & 38 pairs & 3 & -- & 20 & -- & \texttt{gemini 3 flash} \citep{gemini-3-flash} \\
        \bottomrule
    \end{tabular}
\end{table}

For ALE-Bench and MLE-Bench, we use the benchmarks' own recommended \textit{lite} sets. ALE-Bench lite contains 10 of the benchmark's 40 problems, curated by its authors \citep{imajuku2025alebench}. MLE-Bench lite contains 22 of the benchmark's 75 competitions, the split its authors recommend when the full set is too resource-intensive \citep{mlebench}. For FML-Bench, we use the full 18-task benchmark to evaluate agents. KernelBench \citep{ouyang2025kernelbench} is used in \cref{sec:hacking} to study reward hacking. It scores an optimized GPU kernel by timing it in isolation on a fixed test input, a procedure an agent can game by producing a kernel that looks fast on that test but breaks or slows down when used inside a real model. To measure this, we consider kernels that can be used in training pipelines for GPT-2, ViT, or CNN models. For each (kernel, training-context) pair, we compare the isolated-benchmark speedup the agent optimized for against the speedup the kernel actually delivers inside the training loop, each averaged over the three seeds. A kernel counts as reward hacking when its isolated speedup exceeds 1.02$\times$ and either less than half of that speedup survives inside training or the kernel crashes at runtime there.
On WeatherBench~2 \citep{weatherbench2} we treat forecast quality as the objective the agent optimizes. The starting point is a dynamical core and for four headline fields (500\,hPa geopotential, 2\,m temperature, mean sea-level pressure, and 10\,m zonal wind) at lead times from 1 to 15 days, we score the forecast against ERA5 and turn the error into a skill number relative to a persistence forecast. We report the \emph{forecast-skill gain}: how much an agent's optimized dynamical core improves over the unmodified starting core. This measures a relative improvement over its own starting point rather than an absolute score on the WeatherBench~2 leaderboard.

\section{\texorpdfstring{\boldmath\aidehuman{} is a strong baseline}{AIDE human is a strong baseline}}
\label{app:fml}

\begin{figure}[!htbp]
    \centering
    \includegraphics{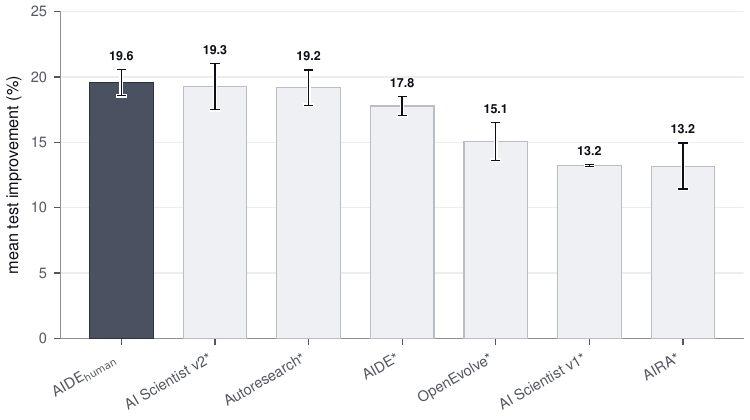}
    \caption{\textbf{\boldmath\aidehuman{} ranks among the strongest code optimization agents on FML-Bench.} \aidehuman{} outperforms all six agents evaluated by the benchmark's authors and represents a strong baseline when comparing agents discovered by \aidesq{}. We show mean normalized test improvement over 18 tasks and 3 seeds with $\pm$1 standard error of the mean across the three seed-level means. Asterisks mark agents whose per-round results were provided by the FML-Bench authors \citep{zou2026fmlbench}.}
    \label{fig:fml-community}
\end{figure}

FML-Bench \citep{zou2026fmlbench} comprises 18 ML research tasks across 10 domains. All agents here are evaluated under the protocol in \cref{tab:transfer-protocol}. \Cref{fig:fml-community} places \aidehuman{} alongside six agents evaluated by the benchmark's authors. We omit AdaptiveSearch \citep{zou2026fmlbench}, the agent introduced with the benchmark itself, because the comparison is meant to measure \aidehuman{} against other strong agents developed independently to this benchmark. \aidehuman{} scores above all six agents, namely AI Scientist v2 \citep{aiscientistv2}, Autoresearch \citep{karpathy2026autoresearch}, the original AIDE agent \citep{aide2025}, OpenEvolve \citep{openevolve}, AI Scientist v1 \citep{aiscientist2026}, and AIRA \citep{airadojo}. \Cref{fig:fml-community} demonstrates that \aidehuman{} is a competitive AI research agent. The margins separating the leading agents are small relative to the seed-level standard error.

\section{Harness gains from recursive self-improvement transfer across models}
\label{app:harness-model}

\begin{figure}[!htbp]
    \centering
    \includegraphics{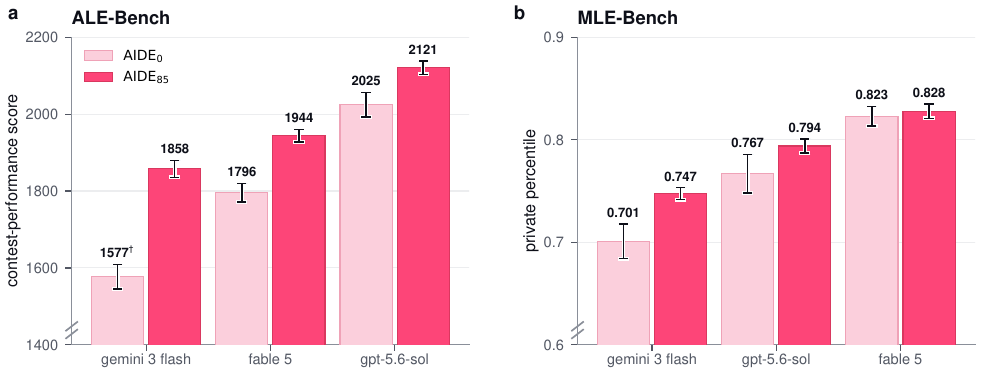}
    \caption{\textbf{Gains from recursive self-improvement transfer across models.} \textbf{a}, Mean private contest performance on ALE-Bench (10 tasks, 10 seeds per task). \textbf{b}, Mean private percentile on MLE-Bench (22 tasks, 3 seeds per task). Bars show benchmark means for \aidek{0} and \aidek{85} with each of three models at a \$20 budget per run. Error bars show $\pm$1 standard error of the benchmark mean. $\dagger$48 runs reached the model's context-window limit and are scored on the best candidate available at termination. The vertical axes are truncated.}
    \label{fig:harness-model}
\end{figure}

We isolate how the harness interacts with the strength of the underlying model by studying agent performance on ALE-Bench and MLE-Bench, using \aidek{0} and \aidek{85}, which effectively mark the start and end of the recursive self-improvement run described in \cref{sec:run}. We evaluate their performance using \texttt{gemini 3 flash}, \texttt{gpt-5.6-sol} \citep{gpt-5-6-sol}, and \texttt{fable 5} \citep{fable-5}. Note that \texttt{gemini 3 flash} is the model used during the selection process while running recursive self-improvement. We follow the protocol summarized in \cref{tab:transfer-protocol}. However, since the latter two models are substantially more expensive, we increase the per-run cost from \$5 to \$20 on both benchmarks and the ALE-Bench time limit from 12 to 36 hours. \Cref{fig:harness-model} shows that \aidek{85}'s gains over \aidek{0} transfer across all three models on both benchmarks. On ALE-Bench, \texttt{gemini 3 flash} with \aidek{85} reaches $1858 \pm 22$ and exceeds \texttt{fable 5} with \aidek{0} at $1796 \pm 24$ under the same budget. On MLE-Bench, the gains are smaller. \aidek{85} with \texttt{fable 5}, the strongest model on this benchmark, stays within one standard error of its \aidek{0} score.

\section{Rejected proposals}
\label{app:rejected-proposals}

From the rejected proposals pooled across three recursive self-improvement runs, \cref{tab:rejected-proposals} reports representative graded instances and maps them to related methods. The labels identify conceptual analogues rather than direct implementations. $\Delta g$ is the candidate's grade relative to the incumbent at that step. Deltas of approximately $-0.004$ to $-0.007$ are small relative to observed run-to-run variability. Out of the graded rewrites that were rejected, about a quarter scored higher than the incumbent on the agent-visible public signal and were rejected on the private grade.

The rejected proposals span a portion of the classical search and learning toolbox, from population-based and restart-driven exploration to robust selection and ensembling. Proposals concentrated most heavily on the search policy and on selection mechanics, followed by context management and robustness changes. Ensembling methods were explored during the recursive self-improvement run but were never retained, with the agent's analyses consistently noting that ensembling LLM calls consumes budget that could otherwise fund additional search steps.

\begin{table}[H]
    \centering
    \footnotesize
    \caption{\textbf{Representative rejected proposals from the recursive self-improvement trajectories.} Scores report $g(a_k)$; deltas report $\Delta g$ relative to the incumbent.}
    \label{tab:rejected-proposals}
    \begin{tabular}{p{0.3\linewidth}p{0.38\linewidth}p{0.24\linewidth}}
        \toprule
        \textbf{What the loop proposed} & \textbf{Related method} &
            \textbf{\boldmath $g(a_k)$ $\cdot$ $\Delta g$} \\
        \midrule
        island populations with periodic migration and crossover & distributed and island-model genetic algorithms \citep{tanese1989distributed,cantupaz1998survey} & 0.685 $\cdot$ $-0.021$ \\
        \addlinespace[2.5pt]
        pairwise LLM-judge tournaments for selection & tournament selection \citep{goldberg1991comparative} and pairwise LLM judging \citep{zheng2023judging} & 0.654 $\cdot$ $-0.090$ \\
        \addlinespace[2.5pt]
        stagnation-triggered exploration escalation & reactive search and adaptive restarts \citep{battiti2008reactive} & 0.704 $\cdot$ $-0.040$ \\
        \addlinespace[2.5pt]
        keep-refining vs.\ restart decisions & restart policies \citep{luby1993optimal} & 0.666 $\cdot$ $-0.078$ \\
        \addlinespace[2.5pt]
        majority-vote ensembles at submission & bagging-style ensemble voting \citep{breiman1996bagging} and self-consistency-style majority voting \citep{wang2023selfconsistency} & up to 0.723 $\cdot$ $-0.031$ (3 variants) \\
        \addlinespace[2.5pt]
        explore-rate tuning and decay schedules & decaying $\epsilon$-greedy exploration \citep{sutton2018reinforcement} & 0.748 $\cdot$ $-0.006$, \emph{within noise} \\
        \addlinespace[2.5pt]
        variance-adaptive exploration boosts & UCB-V \citep{audibert2009exploration} & 0.674 $\cdot$ $-0.080$ \\
        \addlinespace[2.5pt]
        ancestor-descendant trend propagation in selection & MCTS value backup \citep{kocsis2006bandit} & 0.7497 $\cdot$ $-0.004$, \emph{within noise} \\
        \addlinespace[2.5pt]
        promote-second-best and robust near-tie overrides & selection under the optimizer's curse \citep{smith2006optimizers} & 5+ variants $\cdot$ best $-0.007$, \emph{within noise} \\
        \addlinespace[2.5pt]
        treating heavy revisits of one node as overfitting suspicion & no direct analogue identified & 0.7498 $\cdot$ $-0.004$, \emph{within noise} \\
        \bottomrule
    \end{tabular}
\end{table}

\section{Prompt compression}
\label{app:compression}

\begin{figure}[!htbp]
    \centering
    \includegraphics{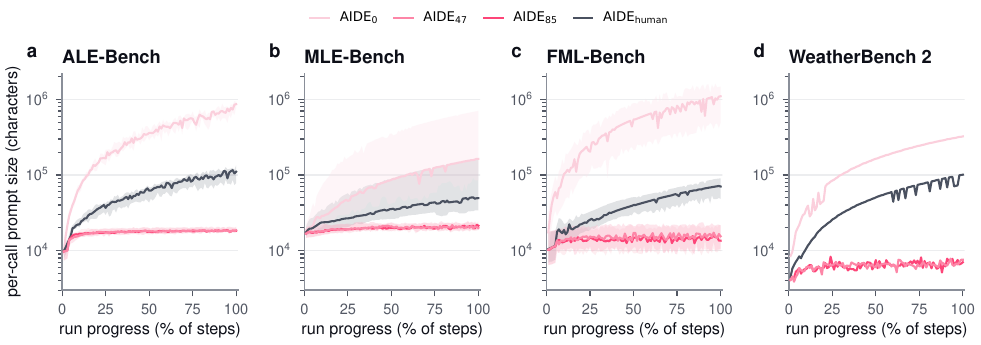}
    \caption{\textbf{\boldmath The discovered agents hold prompts to a bounded size, while \aidek{0}'s prompts grow with run history.} \textbf{a}--\textbf{d}, Per LLM call prompt size in characters (log scale) across every held-out run on ALE-Bench (10 tasks), MLE-Bench (22 tasks), FML-Bench (18 tasks), and WeatherBench~2 (1 task). Lines and bands show the median and interquartile range over the task means. WeatherBench~2's single task carries no band.}
    \label{fig:token-efficiency}
\end{figure}

As discussed in \cref{sec:discoveries}, one of the mechanisms evolved during recursive self-improvement is the context management system. \Cref{fig:token-efficiency} measures its effect on prompt size over the course of a run. For every run of \aidek{0}, \aidek{47}, \aidek{85}, and \aidehuman{} on the held-out benchmarks used in \cref{sec:transfer}, we rebuild the prompt each agent assembled at every step. \aidek{0} concatenates the code and execution output of every prior candidate into each drafting and improvement prompt, so its prompts grow as its run progresses. The discovered agents replace this history with a bounded summary, so their prompts stay small in comparison. This bounded history also removes a concrete failure mode. \aidek{0} terminates when its assembled prompt exceeds the model's context window, a failure that ended five of its FML-Bench runs (\cref{fig:transfer}) and 48 of its ALE-Bench runs at the larger per-run budget in \cref{app:harness-model}. \aidek{85}, \aidek{47} and \aidehuman{} did not encounter any such issues. Since run lengths can differ across agents and tasks, we express progress as a fraction of each run's steps. At each point of run progress, prompt sizes are averaged over seeds within a task. We express this as task-paired prompt reduction factors by dividing \aidek{0}'s task-mean prompt size by the other agents' at the same progress point. This provides us with the compression achieved over \aidek{0}. \Cref{fig:token-efficiency} shows the median and interquartile range of the task means, with \Cref{fig:token-efficiency-main} following the same approach. WeatherBench~2 contributes a single task, so its panels show that task's seed-mean curve without any interquartile band. The median reduction grows throughout the run, reaching 7$\times$ on MLE-Bench, over 40$\times$ on WeatherBench~2, and about 50$\times$ on ALE-Bench and FML-Bench. When measured against \aidehuman{}, the discovered agents' prompts are 2.6--5.7$\times$ smaller on ALE-Bench, MLE-Bench, and FML-Bench and 13--14$\times$ smaller on WeatherBench~2 by the end of the run, while \aidek{0}'s are 3--14$\times$ larger.

\end{document}